\documentclass[10pt,twocolumn,letterpaper]{article}

\usepackage{cvpr}
\usepackage{times}
\usepackage{epsfig}
\usepackage{graphicx}
\usepackage{amsmath}
\usepackage{amssymb}
\usepackage{xcolor}
\usepackage{booktabs}
\usepackage{multicol}
\usepackage{subcaption}
\usepackage{enumitem}
\usepackage{dsfont}
\usepackage{placeins}
\newcommand{\norm}[1]{\left\lVert#1\right\rVert}

\newcommand{\methodname}{DOPS\xspace}

\usepackage[pagebackref=true,breaklinks=true,letterpaper=true,colorlinks,bookmarks=false]{hyperref}

\cvprfinalcopy
\def\cvprPaperID{****}

\begin{document}

\title{\methodname: Learning to Detect 3D Objects and Predict their 3D Shapes}

\author{Mahyar Najibi\textsuperscript{1} \hspace{1em} Guangda Lai\textsuperscript{2} \hspace{1em} Abhijit Kundu\textsuperscript{2} \hspace{1em} Zhichao Lu\textsuperscript{2} \hspace{1em} Vivek Rathod\textsuperscript{2} \\ \hspace{1em} Thomas Funkhouser\textsuperscript{2} \hspace{1em} Caroline Pantofaru\textsuperscript{2} \hspace{1em} David Ross\textsuperscript{2} \hspace{1em} Larry S. Davis\textsuperscript{1} \hspace{1em} Alireza Fathi\textsuperscript{2}\\
\\
{\textsuperscript{1}University of Maryland \hspace{4em} \textsuperscript{2}Google}
}

\maketitle

\begin{abstract}
We propose \methodname, a fast single-stage 3D object detection method for LIDAR data. Previous methods often make domain-specific design decisions, for example projecting points into a bird-eye view image in autonomous driving scenarios. In contrast, we propose a general-purpose method that works on both indoor and outdoor scenes. The core novelty of our method is a fast, single-pass architecture that both detects objects in 3D and estimates their shapes. 3D bounding box parameters are estimated in one pass for every point, aggregated through graph convolutions, and fed into a branch of the network that predicts latent codes representing the shape of each detected object. The latent shape space and shape decoder are learned on a synthetic dataset and then used as supervision for the end-to-end training of the 3D object detection pipeline. Thus our model is able to extract shapes without access to ground-truth shape information in the target dataset. During experiments, we find that our proposed method achieves state-of-the-art results by $\sim$5\% on object detection in ScanNet scenes, and it gets top results by 3.4\% in the Waymo Open Dataset, while reproducing the shapes of detected cars.
\end{abstract}

\section{Introduction}
There has been great progress in recent years on 3D object detection for robotics and autonomous driving applications. Previous work on 3D object detection takes one of these following approaches: (a) projecting LIDAR points to 2D bird's-eye view and performing 2D detection on the projected image, (b) performing 2D detection on images and using a frustum to overlap that with the point cloud, or (c) using a two-stage approach where points are first grouped together and then an object is predicted for each group.  

Each of these approaches come with their own drawbacks. Projecting LIDAR to a bird's-eye view image sacrifices geometric details which may be critical in cluttered indoor environments. The frustum based approaches are strictly dependent on the 2D detector and will miss an object entirely if it is not detected in 2D. Finally, the two-stage methods introduce additional hyperparameters and design choices which require tuning and adapting for each domain separately. Furthermore, we believe grouping the points is a harder task than predicting 3D objects. Solving the former to predict the latter will result in an unnecessary upper-bound that limits the accuracy of 3D object detection.
\begin{figure}
\centering
    \centering
    \includegraphics[width=0.75\linewidth]{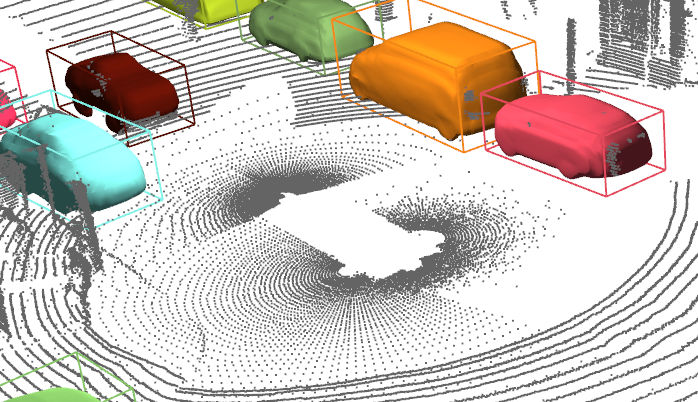}
    \caption{A sample output of our object detection pipeline. }
    \label{fig:teaser}
\end{figure}

In this paper, we propose a single-stage 3D object detection method that outperforms previous approaches. We predict 3D object properties for every point while allowing the information to flow in the 3D adjacency graph of predictions. This way, we do not make hard grouping decisions while at the same time let the information to propagate from each point to its neighborhood. 

In addition to predicting 3D bounding boxes, our pipeline can also output the reconstructed 3D object shapes as depicted in Figure \ref{fig:teaser}. Even though there have been various approaches proposed for predicting 3D bounding boxes, predicting the 3D shapes and extents of objects remains largely under-explored. The main challenges in predicting the 3D shapes of objects are sparsity in LIDAR scans, predominant partial occlusion, and lack of ground-truth 3D shape annotations. In this work, we address these challenges by proposing a novel weakly-supervised approach.

Our proposed solution for shape prediction consists of two steps. First, we learn 3D object shape priors using an external 3D CAD-model dataset by training an encoder that maps an object shape into an embedding representation and a decoder that recovers the 3D shape of an object given its embedding vector. Then, we augment our 3D object detection network to predict a shape embedding for each object such that its corresponding decoded shape best fits the points observed on the surface of that object. Using this as an additional constraint, we train a network that learns to detect objects, predict their semantic labels, and their 3D shapes.

To summarize, our main contributions are as follows. First, we propose a single-stage 3D object detection method that achieves state-of-the-art results on both indoor and outdoor point cloud datasets. While previous methods often make certain design choices (\eg projection to a bird-eye view image) based on the problem domain, we show the possibility of having a generic pipeline that aggregates per-point predictions with graph convolutions. By forming better consensus predictions in an end-to-end hybrid network, our approach outperforms previous works in both indoor and outdoor settings while running at a speed of 12ms per frame. Second, in addition to 3D bounding boxes, our model is also able to jointly predict the 3D shapes of the objects efficiently. Third, we introduce a training approach that does not require ground-truth 3D shape annotations in the target dataset (which is not available in large-scale self-driving car datasets). Instead, our method learns a shape prior from a dataset of CAD models and transfers that knowledge to the real-world self-driving car setup.

\section{Related Works}

\begin{figure*}[!ht]
\centering
\includegraphics[width=\linewidth]{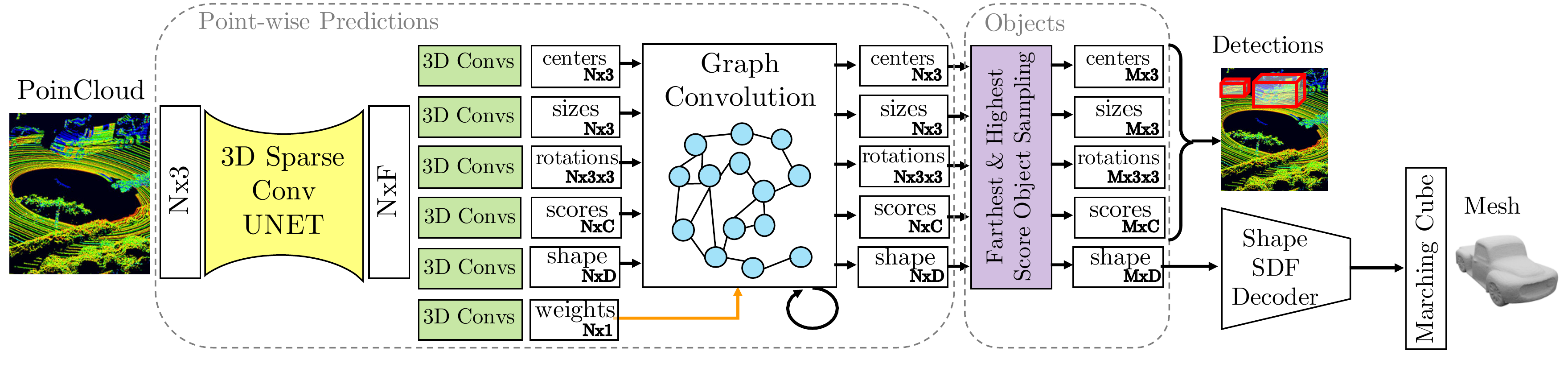}

\caption{Object Detection Pipeline. After voxelization, a 3D sparse U-Net \cite{graham2017submanifold} is used to extract features from each voxel. Then two blocks of sparse convolutions predict object properties per voxel. These features are then propagated back to the points and passed through a graph convolution module. Finally, a ``farthest \& highest score object sampling'' layer followed by NMS outputs the per-object properties including the 3D shape.}
\label{fig:object_detection_pipeline}
\end{figure*}

\subsection{3D Object Detection}
3D object detection has been studied extensively.  In this paper, we focus on applications such as autonomous driving, where the input is a collection of 3D points captured by a LIDAR range sensor. Processing this type of data using neural networks introduces new challenges. Most notably, unlike images, the input is highly sparse, making it inefficient to uniformly process all locations in the 3D space. 

To deal with this problem, PointNet \cite{qi2017pointnet, qi2017pointnet++} directly consumes the 3D coordinates of the sparse points and processes the point cloud as a set of unordered points. FoldingNet \cite{yang2018foldingnet}, AtlasNet \cite{groueix2018atlasnet}, 3D Point Capsule Net \cite{zhao20193d}, and PointWeb \cite{zhao2019pointweb} improve the representation by incorporating the spatial relationships among the points into the encoding process. For the task of 3D object detection, various methods rely on PointNets for processing the point cloud data. To name a few, Frustum PointNets \cite{qi2018frustum} uses these networks for the final refinement of the object proposals and PointRCNN \cite{shi2019pointrcnn} employs PointNets for the task of proposal generation. VoteNet \cite{qi2019deep} deploys PointNet++ to directly predict bounding boxes from points in a two-stage voting scheme.

Projecting the point cloud data to a 2D space and using 2D convolutions is an alternative approach for reducing the computation. Bird's-eye view (BEV), front view, native range view, and learned projections are among such 2D projections. PIXOR \cite{yang2018pixor}, Complex YOLO \cite{simon2018complex}, and Complexer YOLO \cite{simon2019complexer} generate 3D bounding boxes in a single stage based on the projected BEV representation. Chen \etal \cite{chen2017multi}  and Liang \etal \cite{liang2018deep} use a BEV representation and fuse its extracted information with RGB images to improve the detection performance.  VeloFCN \cite{li2016vehicle} projects the points to the front view and uses 2D convolutions for 3D bounding box generation. Recently, LaserNet \cite{meyer2019lasernet} shows that it is possible to achieve state-of-the-art results while processing the more compact native range view representation. PointPillars \cite{lang2019pointpillars}, on the other hand, learns this 2D projection by training a PointNet to summarize the information of points that lie inside vertical pillars in the 3D space.

Voxelization followed by 3D convolutions is also applied to point cloud-based object detection \cite{zhou2018voxelnet}. However, 3D convolution is computationally expensive, especially when the input has a high spatial resolution. Sparse 3D convolution \cite{engelcke2017vote3deep,sparseconvs2015,graham20183d} is shown to be effective in solving this problem. Our backbone in this paper uses voxelization with sparse convolutions to process the point cloud.

Modeling auxiliary tasks is also studied in the literature. Fast and Furious \cite{luo2018fast} performs detection, tracking, and motion forecasting using a single network. HDNET  \cite{yang2018hdnet} estimates high-definition maps from LIDAR sweeps and uses the geometric features to improve 3D detection. Liang \etal \cite{liang2019multi} performs 2D detection, 3D detection, ground estimation, and depth completion. Likewise, our system predicts the 3D shape of the objects from incomplete point clouds besides detecting the objects.

\subsection{3D Shape Prediction for Object Detection}
For 3D object detection from images, 3D-RCNN \cite{kundu20183d} recovers the 3D shape of the objects by estimating the pose of known shapes. A render and compare loss with 2D segmentation annotation is used as supervision. Instead of using known shapes, Mesh R-CNN \cite{gkioxari2019mesh} first predicts a coarse voxelized shape followed by a refinement step. The 3D ground-truth information is assumed to be given. For semantic segmentation, \cite{kuo2019shapemask} improved the generalization of unseen categories by estimating the shape of the detected objects. For 3D detection, GSPN \cite{yi2019gspn} learns a generative model to predict 3D points on objects and uses them for proposal generation. ROI-10D \cite{manhardt2019roi} annotates ground-truth shapes offline and adds a new branch for shape prediction. 
In contrast, our approach does not need 3D shape ground-truth annotations in the target dataset. We use the recently proposed explicit shape modeling \cite{park2019deepsdf,mescheder2019occupancy,saito2019pifu} to learn a function for representing a shape prior. This prior is then used as a weakly supervised signal when training the shape prediction branch on the target dataset.

\section{Approach}
\label{sec:proposed_method}
The overall architecture of our model is depicted in Figure \ref{fig:object_detection_pipeline}. The model consists of four parts: 
The first one consumes a point cloud and predicts per point object attributes and shape embedding. 
The second component builds a graph on top of these per-point predictions and uses graph convolutions to transfer information across the predictions.
The third component proposes the final 3D boxes and their attributes by iteratively sampling high scoring boxes which are farthest from the already selected ones.
Finally, the fourth component decodes the predicted shape embeddings into SDF values which we convert to 3D meshes using the Marching Cubes algorithm \cite{lorensen1987marching}.

\subsection{Per Point 3D Object Prediction}

Given a point cloud of size $N \times I$ consisting of $N$ points with $I$-dimensional input features (\eg positions, colors, intensities, normals), first, a 3D encoder-decoder network predicts 3D object attributes (center, size, rotation matrix, and semantic logits) and the shape embedding for every point.

We use \emph{SparseConvNet}~\cite{graham2017submanifold} as backbone to generate per-point features $\{f_i \in \mathbb{R}^F \}_{i=1}^{N}$. Each of the object attributes and the shape embedding vector are computed by applying two layers of 3D sparse convolutions on the extracted $N \times F$ features.

\textbf{Box Prediction Loss}: We represent a 3D object box by three properties: size (length, width, height), center location ($c_x$, $c_y$, $c_z$), and a 3x3 rotation matrix. Given these predictions, we use a differentiable function to compute the eight 3D corners of each predicted box. We apply a Huber loss on the distance between predicted and the ground-truth corners. The loss will automatically propagate back to the size, center and rotation variables.

To compute the rotation matrix, our network predicts 6 parameters: ($cos_x$, $sin_x$, $cos_y$, $sin_y$, $cos_z$, $sin_z$). Then we formulate the rotation matrix as $R = R_x \times R_y \times R_z$.

The benefit of using this loss in comparison to separate losses for rotation, center, and size is that we do not need to tune the relative scale among multiple losses. Our box corner loss propagates back to all and minimizes the predicted corner errors.
We define the per-point box corner regression loss as
\begin{align}
\mathcal{L}_\text{corner}&(P, G) =  \nonumber\\
&\frac{1}{8 \times \sum_{i=1}^N{\mathds{1}(x_i)}} \sum_{i=1}^N \mathds{1}(x_i) \sum_{j=1}^8  \norm{p_i^{(j)} - g_i^{(j)}}_H 
\end{align}
where $||\cdot||_H$ is the \emph{Huber}-loss (\ie smooth L$_1$-loss), and $\mathds{1}(.)$ is binary function indicating whether a point $x_i$ is on an object surface. $P$ and $G$ are the sets of predicted and ground-truth corners in which $p_i^{(j)}$ represents the $j$'th predicted corner for point $i$, and $g_i^{(j)}$ denotes the corresponding ground-truth corner.  

\textbf{Dynamic Classification Loss}: Every point in the point cloud predicts a 3D bounding box. The box prediction loss forces each point to predict the box that it belongs to. Some of the points make more accurate box predictions than others. Thus we design a classification loss that classifies points that make accurate predictions as positive and others as negative. During the training stage, at each iteration, we compute the IoU overlap between predicted boxes and ground-truth matches and classify the points that have an IoU more than 70\% as positive and the rest as negative. This loss gives us a few percent improvements in comparison to regular classification loss (where we would label points that fall inside an object as positive and points outside as negative). We use a softmax loss for classification.

\subsection{Object Proposal Consolidation}
Each point predicts its object center, size, and rotation matrix. We create a graph where the points are the nodes, and each point is connected to its $K$ nearest neighbors in the center space. In other words, each point is connected to those with similar center predictions. We perform a few layers of graph convolution to consolidate the per-point object predictions. A weight value is estimated per point by the network which determines the significance of the vote a point casts in comparison to its neighbors. We update each object attribute predicted by points as follows:

\begin{equation}
    a_x = \frac{\Sigma_{y \in \mathcal{N}_x} w_y.a_y}{\Sigma_{y \in \mathcal{N}_x} w_y}
\end{equation}

where $a_x$ is an object attribute (\eg object length) predicted for point $x$, $\mathcal{N}_x$ is the set of neighbors of $x$ in the predicted center space, and $w_y$ is the weight predicted for point $y$.

We apply the bounding box prediction loss both before and after the graph convolution step to let the network learn a set of weights that make final predictions more accurate. In this way, instead of directly applying a loss on the predicted point weights, the network automatically learns to assign larger weights to more confident points.

\subsection{Proposing Boxes}
Our network predicts a 3D object box and a semantic score for every point. During the training stage, we apply the losses directly to the per point predictions. However, during the evaluation, we need to use a box proposal mechanism that can reduce the hundreds of thousands of box predictions into a few accurate box proposals. We can greedily pick boxes with high semantic scores. However, we also want to encourage spatial diversity in the locations of the proposed boxes. For this reason, we compute the distance between each predicted box center and all previously selected boxes and choose the one that is far from the already picked points (similar to the heuristic used by KMeans++ initialization \cite{arthur2007kmeanspp}).  More precisely, at step $t$, given predicted boxes for previous steps $\mathcal{B}_{1:t-1}$,  we select a seed point as follows:

\begin{equation*}
    b_t = \arg \max_{b \not\in \mathcal{B}_{1:t-1}} [\log(s_b) + \alpha \log(D(b, \mathcal{B}_{1:t-1})]
\end{equation*}
where
\begin{equation*}
    D(b, \mathcal{B}_{1:t-1}) = \min_{b^\prime \in \mathcal{B}_{1:t-1}} \norm{b - {b^\prime}}
\end{equation*}
and $s_b$ represents the foreground semantic score of box $b$. Selecting boxes with high foreground semantic score guarantees high precision, and selecting diverse boxes guarantees high recall. Note that our sampling strategy is different from the non-maximum suppression algorithm. In NMS, boxes that have a high IoU are suppressed and are not redeemable, while in our algorithm, we can tune the balance between confidence and diversity.

\subsection{Shape Prediction}
\label{sec:shape_prediction}

To predict shapes, first, we learn a shape prior function from an external synthetic 3D dataset of CAD models as discussed in Section \ref{sec:shape_prior}. Then we deploy our learned prior to recover 3D shapes from the embedding space predicted by the object detection pipeline. 

\begin{figure*}[!ht]
\centering
\includegraphics[width=\linewidth]{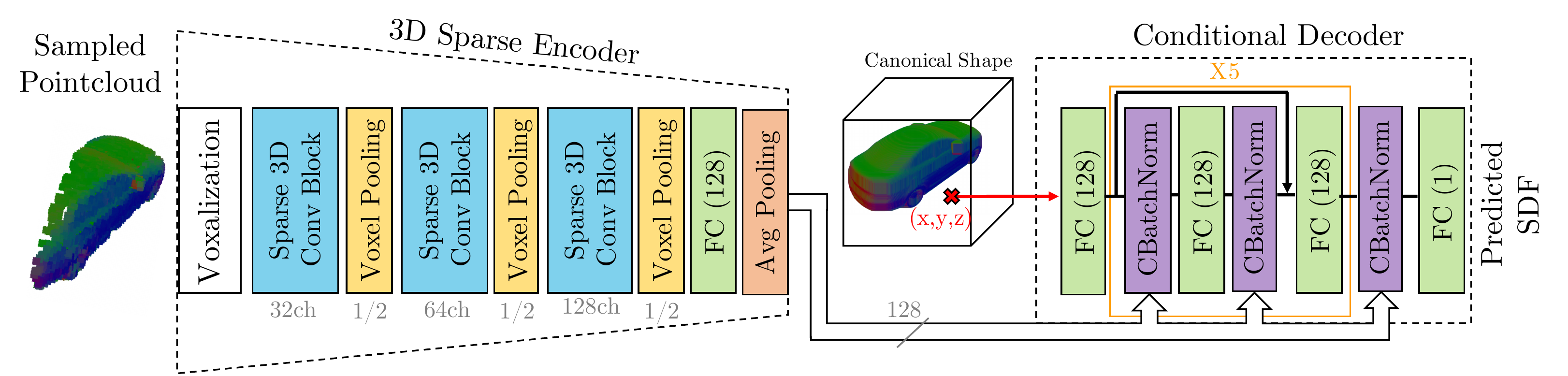}
\caption{Shape Prior Network Architecture. The encoder consumes the point cloud representation of an object after augmentations (\eg random cropping) and outputs a compact embedding vector. The decoder consists of Conditional Batch Norm \cite{de2017modulating} layers which are conditioned on the predicted embeddings. The input to the decoder is a batch of 3D point coordinates and the output is the predicted signed distance of each point to the object surface.}
\label{fig:shape_prior}
\end{figure*}

\subsubsection{Modeling the Shape Prior}
\label{sec:shape_prior}

There are various ways to represent a shape prior. For our application, given that a shape embedding vector should be predicted for each point in the point cloud, the representation needs to be compact. We use an encoder-decoder architecture with a compact bottleneck to model the shape prior. The general framework is depicted in Figure \ref{fig:shape_prior}.

The shape encoder consumes the point cloud of an object after data augmentation techniques (\eg random cropping) and then outputs a compact shape embedding vector. The point cloud representation of the object is first voxelized and then forwarded through an encoder network. The network consists of three convolutional blocks, each having two 3D sparse convolution layers intervened by \textit{BatchNorm} and \textit{ReLU} layers (not shown in the figure for simplicity). The spatial resolution of the feature maps is reduced by a factor of two after each convolutional block. Finally, a fully-connected layer followed by a global average pooling layer output the embedding vector of the input shape.

For shape decoding, we represent the shape as a level set of an implicit function \cite{mescheder2019occupancy, saito2019pifu, park2019deepsdf}. That is, the shape is modeled as the level set zero of a signed distance field (SDF) function over a unit hyper-cube. Following \cite{mescheder2019occupancy}, we rely on Conditional Batch Normalization\cite{de2017modulating, dumoulin2016adversarially} layers to condition the decoder on the predicted embedding vector. The input to the decoder is a batch of 3D coordinates of the query points. After five conditional blocks, a fully connected layer followed by a \textit{tanh} function predicts the signed distance of each query from the surface of the object in a canonical viewpoint.

During training, we sample some query points close to the object surface and some uniformly in the unit hypercube surrounding the object to predict their SDF values. However, as suggested in \cite{saito2019pifu}, we regress towards discrete label values to capture more detail near the surface boundaries. More precisely, given a batch of training queries $Q = \{q_i\}_{i=1}^N \in \mathbb{R}^{3\times N}$, their corresponding ground-truth signed distance values $S = \{s_i\}_{i=1}^N \in \mathbb{R}^{N}$, and their predicted embedding vectors $E = \{e_i\}_{i=1}^N \in \mathbb{R}^{D\times N}$, the loss is defined as:
\begin{equation}
    \mathcal{L}(Q,S,E | f) = \frac{1}{N} \sum_{i=1}^N{\norm{f(q_i | e_i) - sign(s_i)}}^2
    \label{eq:shape_loss}
\end{equation}
where $f(.)$ is the conditional decoder function and $sign(.)$ is the sign function.

\subsubsection{Training the Shape Prediction Branch}
Although there is no ground-truth 3D shape annotation available in detection datasets collected for applications such as autonomous driving, once trained, the learned prior model can be deployed to enforce shape constraints. That is, for each object in the incomplete point cloud, we expect that most of the observed points in its bounding box lie on its surface.

To predict the shape embedding, we add a branch to the object detection pipeline to predict a $D$-dimensional vector per point. The shape embeddings for all points belonging to an object is then averaged pool to form its shape representation. To enforce the constraints, we freeze the 3D decoder in Figure \ref{fig:shape_prior} and discard the encoder. Conditioned on the predicted shape embedding and given some shape queries per object, the frozen shape decoder should be able to predict the signed distances. 

To define the queries, for each object present in the point cloud, we subtract the object center and rotate the queries to match the canonical orientation used during training the shape prior network. Then, the queries are projected into a unit hyper-cube. We also pre-process them by removing points on the ground and augmenting the symmetrical points (if the object is symmetrical). Finally, as the shape prior is trained with discrete sign labels, we sample some number of queries on the ray connecting the object center to each of the observed points and assign -1/+1 labels to inside/outside queries respectively (in this paper we sample two points with distance $\delta=0.1$ to each observed point along the rays). During training, we also optimize the loss defined in Eq. \ref{eq:shape_loss} for objects with a reasonable number of points observed (\ie minimum of 500 points in this paper.)

\subsection{Achieving Real-Time Speed}
Our 3D sparse feature extractor with $30$ 3D sparse convolution layers, $7$ 3D sparse pooling layers, and $7$ 3D sparse un-pooling layers achieves a speed of $12ms$ per frame on Waymo Open dataset (with around 200k input points per frame). Here we describe the implementation details of our Tensorflow sparse GPU operations.

We use CUDA to implement the submanifold sparse convolution \cite{graham2017submanifold} and sparse pooling GPU operations in TensorFlow. Since the input to the convolution operation is sparse, we need a mechanism to get all the neighbors for each non-empty voxel. We implemented a hashmap to do that, where the keys are the XYZ indices of the voxels, and the values are the indices of the corresponding voxels in the input voxel array. We use an optimized spatial hash function\cite{teschner2003optimized}. Our experiments on the Waymo Open dataset shows that with a load factor of $0.42$, the average collision rate is $0.18$. We precompute the neighbor indices for all non-empty voxels and reuse them in one or more subsequent convolution operations. We use various CUDA techniques to speed up the computation (\eg partitioning and caching the filter in shared memory and using bit operations).

Both 3D sparse max pooling and 3D sparse average pooling operations are implemented in CUDA. Since each voxel needs to be looked up only once during pooling, we do not reuse the convolution hashmap that can introduce redundant lookups. Instead, we compute the pooled XYZ indices and use them as the key to building a new ``hashmultimap''(multiple voxels can be pooled together thus having the same key), and shuffle the voxels based on the keys. Our experiments show that this approach is more than 10X faster than the radix sort provided by the CUB library. Furthermore, since our pooling operation does not rely on the original XYZ indices, it has the ability to handle duplicate input indices. This allows us to use the same operation for voxelizing the point cloud, which is the most expensive pooling operation in the network. Our implementation is around 20X faster than a well-designed implementation with pre-existing TensorFlow operations.

\section{Experiments}
\subsection{Experimental Setup}
For our object detection backbone, we use an encoder-decoder UNET with sparse 3D convolutions. The encoder consists of 6 blocks of 3D sparse convolutions, each of which having two 3D sparse convolutions inside. Going deeper, we increase the number of channels gradually (\ie 64, 96, 128, 160, 192, 224, 256 channels).  We also apply a 3D sparse pooling operation after each block to reduce the spatial resolution of the feature maps. For the decoder, we use the same structure but in the reverse order and replace the 3D sparse pooling layers with unpooling operations.  Two 3D sparse convolutions with 256 channels connect the encoder and decoder and form the bottleneck. Models are trained on 20 GPUs with a batch size of 6 scenes per each. We use stochastic gradient descent with an initial learning rate of 0.3 and drop the learning rate every 10K iterations by the factors of [1.0, 0.3, 0.1, 0.01, 0.001, 0.0001]. We use a weight decay of $5 \times 10^{-4}$ and stop training when the loss plateaus. We use random rotations of (-10, 10) degrees along the z-axis and random scaling of (0.9, 1.1) for data augmentation.

The 3D sparse encoder in our shape prior network consists of three convolutional blocks with two 3D sparse convolutions in each. We use an embedding size of 128 dimensions and set $((32, 64), (64, 128), (128,128))$ as the number of channels in the 3D convolutional layers. We down-sample the feature maps by a factor of 2 after each block. A global average pooling, followed by a fully-connected layer outputs the predicted embedding. Our shape decoder consists of five conditional blocks with two 128 dimensional fully connected layers intervened by conditional batch normalization layers. A \emph{tanh} function maps predictions to [-1, +1]. We train our model with an initial learning rate of 0.1 with the same step-wise learning rate schedule used for training the detection pipeline.

\subsection{Datasets}
\textbf{ScanNetV2} \cite{Dai17scannet} is a dataset of 3D reconstructed meshes of around 1.5K indoor scenes with both 3D instance and semantic segmentation annotations. The meshes are reconstructed from RGB-D videos that are captured in various indoor environments. Following the setup in \cite{qi2019deep}, we sample
vertices from the reconstructed meshes as our input point clouds and since ScanNetV2 does not provide amodal or oriented bounding box annotations, we predict axis-aligned bounding boxes instead, as in \cite{qi2019deep, hou2019sis}.

\textbf{Waymo Open Dataset} \cite{ngiam2019starnet, zhou2019multiview} is a large scale self-driving car dataset, recently released for benchmarking 3D object detection. The dataset captures multiple major cities in the U.S., under a variety of weather conditions and across different times of the day. The dataset contains a total of 1000 sequences, where each sequence consists of around 200 frames that are 100 ms apart. The training split consists of 798 sequences containing 4.81M vehicle boxes. The validation split consists of 202 sequences with the same duration and sampling frequency, containing 1.25M vehicle boxes. The effective annotation radius in the Waymo Open dataset is 75m for all object classes. For our experiments, we evaluate 3D object detection metrics for vehicles and predict 3D shapes for them.

\begin{table}[t!]
\small
\setlength{\tabcolsep}{3.2pt}
\begin{center}
\resizebox{\columnwidth}{!}{  
\begin{tabular}{l |c| c c}
    \toprule
    & Input & mAP@0.25 & mAP@0.5 \\ \hline    
    DSS~\cite{song2016deep,hou2019sis} & Geo + RGB & 15.2 & 6.8  \\
    MRCNN 2D-3D~\cite{he2017mask,hou2019sis} & Geo + RGB & 17.3 & 10.5 \\
    F-PointNet~\cite{qi2018frustum,hou2019sis} & Geo + RGB & 19.8 & 10.8 \\
    GSPN~\cite{yi2018gspn} & Geo + RGB & 30.6 & 17.7 \\ \midrule
    3D-SIS \cite{hou2019sis} & Geo + 1 view & 35.1 & 18.7 \\ 
    3D-SIS \cite{hou2019sis} & Geo + 3 views & 36.6 & 19.0 \\
    3D-SIS \cite{hou2019sis} & Geo + 5 views & 40.2 & 22.5 \\ \midrule
    3D-SIS \cite{hou2019sis} & Geo only & 25.4 & 14.6 \\
    DeepVote\cite{qi2019deep} & Geo only & 58.6 & 33.5 \\ \midrule
    \methodname (ours) & Geo only & \textbf{63.7} & \textbf{38.2}\\ \bottomrule
\end{tabular}
}
\end{center}
\caption{3D object detection results on ScanNetV2 validation set. We report results for other approaches as appeared in the original papers or provided by the authors.} 
\label{tab:scannet}
\end{table}

\subsection{Object Detection on ScanNetV2}
\begin{figure*}
\centering
    \centering
    \includegraphics[width=0.95\linewidth]{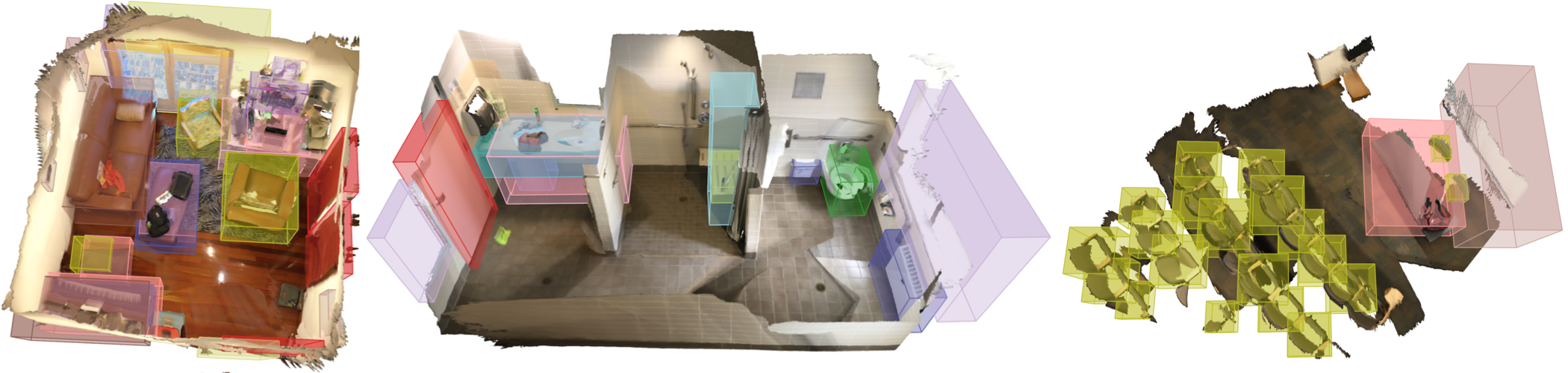}
    \caption{Qualitative results for the axis aligned object dtection on the ScanNet dataset.}
    \label{fig:scannet_quals}
\end{figure*}

We present our object detection results on the ScanNetV2 dataset in Table \ref{tab:scannet}. For this dataset, we follow \cite{qi2019deep, hou2019sis} and predict axis-aligned bounding boxes. Although we only use the available geometric information, we also compare the proposed method with approaches that use the available RGB images and different viewpoints. Our approach noticeably improves the state-of-the-art by 3\% and 4.6\% with respect to mAP@0.25 and mAP@0.5 metrics. We also report our per-category results on the ScanNetV2 dataset in Table \ref{tab:scannet_per_categry}. Figure \ref{fig:scannet_quals} shows our qualitative results.

\begin{table*}[h!]
\centering
\small
\setlength\tabcolsep{1.5pt} 
\resizebox{\textwidth}{!}{\begin{tabular}{c|c|c|c|c|c|c|c|c|c|c|c|c|c|c|c|c|c|c|c}
&Bathtub& Bed& \begin{tabular}{@{}c@{}}Book \\ Shelf\end{tabular} & Cabinet& Chair& Counter& Curtain& Desk& Door & Other & Picture& Refrig. & \begin{tabular}{@{}c@{}}Shower \\ Curtain\end{tabular}&Sink& Sofa & Table & Toilet& Window & \begin{tabular}{@{}c@{}}Overall \\ Score\end{tabular}\\
\toprule
mAP@0.25&86.6 & 83.3 & 41.0& 53.2 & 91.6 & 51.9 & 53.9 & 73.7 & 54.8 & 59.2 & 26.3 & 49.2 & 64.7 & 71.3 & 82.6 & 60.5 & 98.0 & 45.2 & 63.7\\
\toprule
mAP@0.5&71.0 & 70.2 & 21.4& 25.2 & 75.8 & 9.5 & 24.4 & 39.4 & 27.8 & 35.0 & 12.3 & 33.7 & 17.3 & 35.7 & 54.8 & 41.2 & 80.6 & 12.1 & 38.2\\
\end{tabular}}
\caption{Per-category results on ScanNetV2. We report mAP at IoU of $25\%$ and $50\%$.}
\label{tab:scannet_per_categry}
\end{table*}

\subsection{Object Detection on Waymo Open}

We achieve an mAP of \textbf{56.4}\% at IOU of 0.7. This is while StarNet \cite{ngiam2019starnet} achieves an mAP of 53.0\%.  
Note that \cite{zhou2019multiview} also reports 3D object detection results on the Waymo open dataset. However, their results are not directly comparable to ours since they fuse 2D networks applied to multiple views in addition to a 3D network. Since our detection pipeline consists of different parts, we also perform our ablation studies on this dataset. Table \ref{tab:waymo_detection_ablations} shows the contribution of each component of the system on its overall performance. Each column shows the performance when a single component of the system is excluded and the rest remain the same. Removing graph convolution over the predictions on the neighborhood graph reduces the detection performance by $\sim$ 2\%, showing its importance. Replacing the dynamic classification loss with a regular classification loss drops the performance by 3.3\%. Finally, if instead of the farthest and highest object sampling, one directly deploys NMS to form the objects, the performance drops by 0.7\%. We also noticed that shape prediction does not have a noticeable impact on the detection precision. We believe the main reason is that the Waymo Open dataset has manually labeled bounding boxes for object detection, but no ground-truth shape annotations. As a result, the shape predictions are supervised only with noisy, partial, and sparse LIDAR data, which provides a relatively weaker training signal.

\begin{table}[h!]
\centering
\small
\setlength\tabcolsep{1pt} 
\begin{tabular}{c|c|c|c|c}
&  \begin{tabular}{@{}c@{}} \methodname \\ (ours)\end{tabular} &\begin{tabular}{@{}c@{}}w\slash o Graph\\Convolution \end{tabular}& \begin{tabular}{@{}c@{}} w\slash o Dynamic \\ Cls Loss\end{tabular} &  \begin{tabular}{@{}c@{}} w\slash o Farthest \&\\Highest Sampling\end{tabular} \\
\toprule
mAP@0.7& \textbf{56.4} & 54.5 & 53.1 &55.7
\end{tabular}
\caption{The contribution of each compontent on the overal accuracy on the Waymo Open Dataset.}
\label{tab:waymo_detection_ablations}
\end{table}

\begin{figure}[!ht]
\centering
\vspace*{-2mm}
\includegraphics[width=0.75\linewidth]{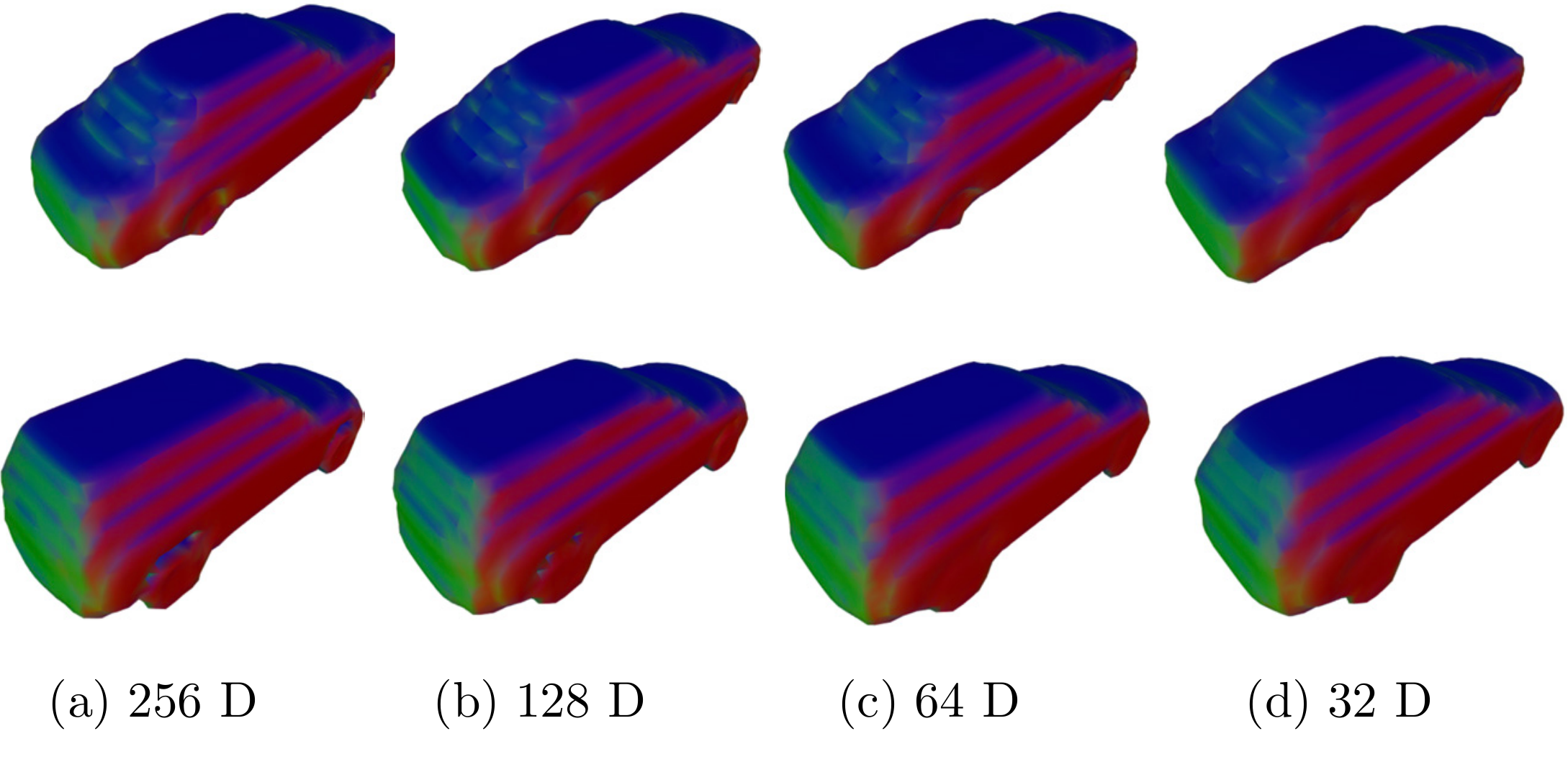}
\caption{Shapes recovered in ShapeNet dataset from the learned embedding by Marching Cube \cite{lorensen1987marching} on a $100^3$ SDF volume. Our prior network captures the shape information even using a low-dimensional embedding vector.}
\label{fig:shape_prior_dimension}
\end{figure}

\subsection{3D Shape Prediction on Waymo Open}

To model shape, we first learn a prior from the synthetic ShapeNet dataset \cite{Chang15shapenet}. Figure \ref{fig:shape_prior_dimension} shows shapes recovered from the compact embedding vectors predicted for CAD models in ShapeNet. Each row represents one shape and columns show the results for different embedding dimensions. We use marching cube \cite{lorensen1987marching} with a resolution of 100 points per side on SDF values predicted by our decoder for a uniform hyper-cube surrounding the object. As can be seen, the decoder can recover the extent of the object from the predicted embedding vector, even when the dimensionality of the embedding space is low.

\begin{figure}[!t]
\centering
\includegraphics[width=\linewidth]{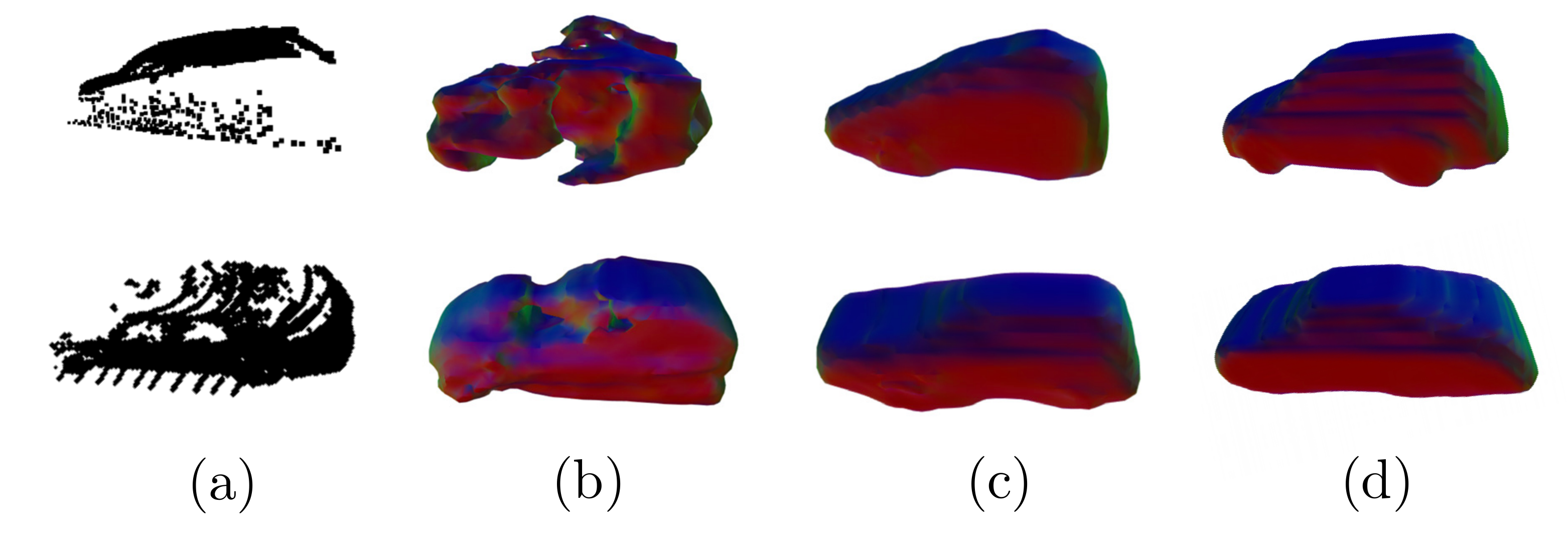}
\caption{Ablations for shape fitting on observed points in Waymo dataset. (a) Observed points. (b) Enforcing the decoder to predict zero SDFs only for the observed points. (c) Adding two points along the rays passing through the observations and object center inside/outside the object with a distance of $\delta=0.5$. (d) Decreasing $\delta$ to $0.1$.}
\label{fig:shape_prior_exp}
\end{figure}

Once trained on the ShapeNet dataset, we freeze the decoder and use it to recover shapes from the observed points in the real-world scenes captured by the LIDAR sensors. However, compared to the synthetic CAD models, LIDAR points are incomplete, noisy, and the distribution of the observed points can be different from the clean synthetic datasets. Consequently, we found proper pre-processing and data augmentation techniques crucial. Noticeably, ShapeNet contains dense annotations even for surfaces inside the objects. However, when it comes to autonomous driving datasets, only a sparse set of points on the surface of the object is observed. We remove internal points when training on the ShapeNet dataset and empirically noticed that this step improves convergence and shape prediction quality. Moreover, the LIDAR sensor frequently captures points on the ground while this does not happen in ShapeNet. We also remove points on the ground based on the coordinate frame of each object.

\begin{figure*}
\centering
    \centering
    \includegraphics[width=\linewidth]{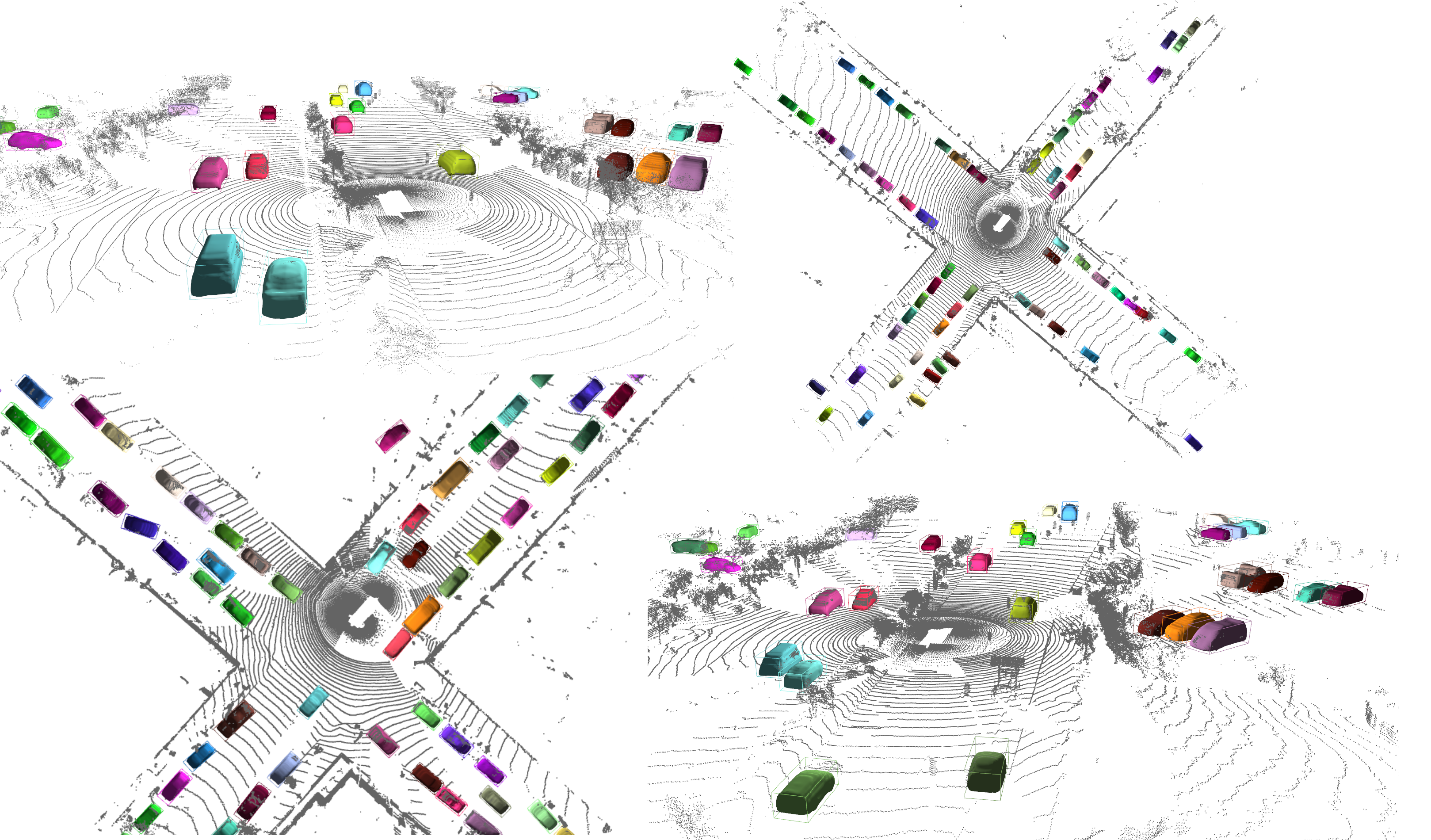}
    \caption{Qualitative results of 3D object detection and 3D shape prediction.}
    \label{fig:open_waymo_quals}
\end{figure*}

Given a set of $N$ observed points in the point cloud, a predicted encoding vector, and a frozen decoder, it is possible to enforce $N$ weakly supervised constraints to recover the shapes. The points which are observed should lie on the surface of the object with a high probability. That is, the frozen decoder conditioned on the predicted embedding should predict a zero SDF value for these points. However, this set of constraints is not enough for reliably recovering the shape. Figure \ref{fig:shape_prior_exp}{\color{red} b} shows the case when a shape is fitted to a set of of points observed from an object in the Waymo Open dataset, shown in \ref{fig:shape_prior_exp}{\color{red} a}. As can be seen, the decoder is able to fit a complex surface to the points. This is while the shape almost perfectly passes through the observed points.

Instead, we augment points with additional ones sampled along the ray connecting the observations to the object center. For each observed point, we add two points on this ray inside and outside the object with distance $\delta$ from the surface and assign labels -1/+1 to them respectively. Figures \ref{fig:shape_prior_exp}{\color{red} c}, and \ref{fig:shape_prior_exp}{\color{red} d} show the shape fitting when we set $\delta$ to $0.5$ and $0.1$ respectively. As can be seen, this augmentation technique is crucial and sampling closer points increases the quality of the recovered shape. 

Finally, Figure \ref{fig:open_waymo_quals} presents our end-to-end shape prediction results. Note that the car shapes fit the point cloud and are not simply copies of examples from a database.

\section{Conclusions}
  We propose \methodname, a single-stage object detection system which operates on point cloud data. \methodname directly predicts object properties for each point. Instead of grouping points before prediction, a graph convolution module is deployed to aggregate the information across neighboring points. For a more accurate localization, it also outputs a 3D mesh using a shape prior learned on a synthetic dataset of CAD models. We show state-of-the-art results for on 3D object detection datasets for both indoor and outdoor scenes.  Topics for future work include detection and tracking over time, semi-supervised training of shape priors, and extending shape models to handle non-rigid objects. 
\FloatBarrier
{\small
\bibliographystyle{ieee_fullname}
\bibliography{egbib}
}

\end{document}